\documentclass[runningheads,a4paper]{llncs}

\usepackage{amssymb}
\usepackage{romannum}
\usepackage{authblk}
\usepackage{graphicx}
\usepackage{multicol}
\usepackage{float}

\usepackage{url}
\urldef{\mailsa}\path|{alfred.hofmann, ursula.barth, ingrid.haas, frank.holzwarth,|
\urldef{\mailsb}\path|anna.kramer, leonie.kunz, christine.reiss, nicole.sator,|
\urldef{\mailsc}\path|erika.siebert-cole, peter.strasser, lncs}@springer.com|    
\newcommand{\keywords}[1]{\par\addvspace\baselineskip
\noindent\keywordname\enspace\ignorespaces#1}

\begin{document}

\mainmatter  

\title{Nuclei Detection Using Mixture Density Networks}

\titlerunning{}

%
%

\author[1,4]{Navid Alemi Koohababni}
\author[2]{Mostafa Jahanifar}
\author[3]{Ali Gooya}
\author[1,4]{Nasir Rajpoot}

\affil[1]{Department of Computer Science, University of Warwick, Coventry}
\affil[2]{Department of Biomedical Engineering, Tarbiat Modares University }
\affil[3]{Department of Electronic and Electrical Engineering, University of Sheffield}
\affil[4]{Alan Turing Institute, London, UK}

\authorrunning{Alemi et al.}

\institute{}

%
%

\toctitle{Lecture Notes in Computer Science}
\tocauthor{Authors' Instructions}
\maketitle

\vspace{-5mm}
\begin{abstract}
Nuclei detection is an important task in the histology domain as it is a main step toward further analysis such as cell counting, cell segmentation, study of cell connections, etc. This is a challenging task due to complex texture of histology image, variation in shape, and touching cells.
To tackle these hurdles,  many approaches have been proposed in the literature where deep learning methods stand on top in terms of performance. Hence, in this paper, we propose a novel framework for nuclei detection based on Mixture Density Networks (MDNs). These networks are suitable to map a single input to several possible outputs and we utilize this property to detect multiple seeds in a single image patch. A new modified form of a cost function is proposed for training and handling patches with missing nuclei. The probability maps of the nuclei in the individual patches are next combined to generate the final image-wide result. The experimental results show the  state-of-the-art performance on complex colorectal adenocarcinoma dataset.

\end{abstract}
\vspace*{-5mm}
\section{Introduction}
\label{sec:intro}
Precise localizing the nucleus in histology images is a main step for successive medical image analysis such as cell segmentation, counting and morphological analysis \cite{grau2004improved}. Unfortunately, robust cell detection  is a challenging task due to nucleus clutters, large variation in shape and texture, nuclear pleomorphism, touching cells and poor image quality \cite{quelhas2010cell}. Since manual detection of nucleus for further diagnostic assessment  imposes a high workload on pathologists,   computer assisted methods have attracted a lot of interest in recent years \cite{schmitt2008radial}. To this end,  many automatic cell detection algorithms are proposed in literature. Parvin et al. \cite{parvin2007iterative} introduced the iterative voting methods which use oriented kernels to localize cell centers, where  the voting direction and areas were updated in each iteration. 
Qi et al \cite{qi2012robust} utilize a single path voting mechanism that is followed by clustering step. Similarly, Hafiane et al \cite{hafiane2008fuzzy} detect the nuclei by clustering the segmented centers using an iterative voting algorithm. 
multiscale Laplacian-of-Gaussian (LOG) \cite{akakin2012automated} and construction of concave vertex graph \cite{yang2008automatic} can also be found in the literature. A popular approach to handle touching cells is based on the watershed algorithm
\cite{grau2004improved,jung2010segmenting}.
However, due to the large variations
in microscopy modality, nucleus morphology, and the inhomogeneous background,
it remains to be a challenging topic for these non-learning methods.
Data-driven methods utilizing hand-crafted features have also been extensively applied for cell detection due to
their promising performance. Interested readers are  referred to \cite{thomas2017review} for more details about methods which rely on hand crafted features and classic supervised methods.

Deep learning has shown an outstanding performance in computer vision  analysis of both natural and biomedical images.
Deep learning methods extract the appropriate features from an image without the need for laborious feature engineering and parameter tunning. Ciresan et al. \cite{cirecsan2013mitosis} applied a deep neural network (DNN) as
a pixel classifier to differentiate between mitotic and non-mitotic nuclei in breast cancer histopathology images. Xie et al. \cite{xie2018efficient} proposed a structured regression convolution neural network (CNN) for nuclei detection wherein the gaussian distribution is fitted on the nucleus center to construct the probability map which is considered as an image mask, then a weighted mean squared loss is minimized via pixel-wise back-propagation. Xu et al. \cite{xu2016stacked}  proposed a stacked sparse autoencoder
strategy to learn high level features from patches of breast histopathology images and then classify these patches as nuclear or non-nuclear.
Sirinukunwattana et al. \cite{sirinukunwattana2016locality} proposed a  locality sensitive deep learning approach for nuclei detection in the H\&E stained colorectal adenocarcinoma histology images. In this approach, a spatially constrained CNN is
first employed to generate a probability map for a given input image using local information.Then the centroids of nucleus are detected by identifying local maximum intensities. 

In this paper, we propose a simple yet effective method based on Mixture Density Networks (MDN) introduced by Bishop \cite{bishop1994mixture} for solving inverse problems, where we have multiple targets for an individual input. MDN learns the distribution of nucleus within an image hypothesizing that each nuclei has a Gaussian distribution with a maximum value on its center.  Here we formalize the concept of MDN for cell detection problem. Due to MDN's flexibility to localize nucleus, we  show that it has a better performance when compared with the  other cell detection algorithms on a challenging colon cancer dataset.

Our contributions in this paper are the followings: \romannum{1})
We define the problem of nuclei detection as mapping a single input image patch into the probability density function (pdf) of the nuclei center, from which the observed locations have been sampled. The pdf is modeled as a Gaussian Mixture Model (GMM) and its parameters are learned via a back-propagation. In addition, a Bernoulli distribution is trained whose parameter predicts if the local patch contains any nucleus and thus the fit of the GMM is liable. 
\romannum{2}) We show the network can detect the nuclei even when trained with a sparse annotated samples, whereas using other methods result in a poor performance. 
\romannum{3}) we demonstrate the capability of algorithm to learn the distribution of nuclei center from the training data without the need to define  fixed variance size for all nucleus as some methods do \cite{sirinukunwattana2016locality,xie2018efficient}.

The rest of this paper is organized as follows: A brief review
of MDN and its generalization to our problem is presented in Sec. \ref{MDN}.   The experimental results and comparison to the state of-the-art
are described and discussed in Sec. \ref{Experimental}, and finally,
some concluding remarks are drawn in Sec. \ref{conclusion}.

\vspace{-5mm}
\section{Mixture Density Networks}\label{MDN}

For a general task of supervised learning our goal is to model a conditional distribution  $p(t|x)$ (for image patch x and nucleus center t),   which is considered Gaussian for many problems and a least square energy function is often obtained using maximum likelihood. These assumptions can lead to a poor performance in many application having plausible non-Gaussian distributions. One of such applications is one to many mapping where one input corresponds to several outputs. The assumption of having a Gaussian posterior distribution forces the model to predict only one output discarding other target values at best. Moreover, The  network prediction is the average of all target values which is incorrect \cite{bishop1994mixture}. To address this problem, we can consider a general framework for modeling the conditional posterior probability distribution by modeling it as a mixture density represented as a linear combination of kernel functions:
\begin{equation}
p(t|x) = \sum\limits_{k = 1}^K {{\alpha _k}(x){\phi _k}(t|x)} 
\end{equation}
where $K$ is the number of components in the mixture and ${{\alpha _i}}$s are mixing coefficients. We assume that kernel functions ${{\phi}(t|x)}$ are isotropic Gaussian:
\begin{equation} \label{Gaussian}
{\phi _k}(t|x) = \frac{1}{{{{(2\pi )}^{{c \mathord{\left/
 {\vphantom {c 2}} \right.
 \kern-\nulldelimiterspace} 2}}}\sigma _k^c(x)}}\exp \left\{ { - \frac{{{{\left\| {t - {\mu _k}(x)} \right\|}^2}}}{{2\sigma _k^2(x)}}} \right\}
\end{equation}
where ${{\mu _k}(x)}$ and ${{\sigma _k}^2(x)}$ are the mean and  the variance of the $k$th Gaussian, respectively, and $c$ is the dimension of target variable.

Here, the parameters of the mixture model are considered to be functions of input image patch $x$. This can be achieved by using a conventional neural network as a function that takes $x$ as input. These layers are then combined with other fully connected layers to from the Mixture Density Network (MDN), (see Fig.\ref{schematic}). Building the MDN increases the number of parameters from c output to $(c + 2) \times K$.
There are some restrictions on these parameters that can be found in detail in \cite{bishop1994mixture}.

To define the error function, the standard negative logarithm of the maximum likelihood is used. Therefore the original loss function for the network is \cite{bishop1994mixture}:

\begin{equation} \label{Gaussian}
E(W) =  - \sum\limits_{n = 1}^N {\ln p({t_n}|{x_n}) =  - \sum\limits_{n = 1}^N {\ln \left( {\sum\limits_{k = 1}^K {{\alpha _k}({x_n})\phi ({t_n}|{x_n})} } \right)} } 
\end{equation}
where  summation over $n$ applies to all dataset. In the next section, we modify this cost function so that it becomes more suitable to handle image patches with multiple and/or missing nuclei. 

\subsection{Extending MDN for Nuclei Detection}
For nuclei detection, deep learning approaches are either provided with small patches each containing one nuclei\cite{sirinukunwattana2016locality,xu2016stacked} or designed as pixel wise structured logistic regression \cite{xie2018efficient,xie2018microscopy}.
Here, we formulate the cell detection as the problem of mapping one to many outputs, as each input vector (image) can have multiple variables defined as  the locations (coordinates) of the nuclei.

To adjust the MDN for nuclei detection, we modify the equation (\ref{Gaussian}) to take one input (image patch) and all of its corresponding target coordinates of the nuclei during training.
 This equation  can only be used when all input patches contain nuclei (when we have at least one target variable for each image), whereas there are many patches with no nucleus. To address this problem, we add a Bernoulli variable  to our loss function to ignore mixture parameters:
\begin{equation} \label{Gaussian1}
\begin{array}{r}
E(W) =  - \sum\limits_{i = 1}^I {\sum\limits_{n = 1}^{{N_i}} {\ln \left\{ {\sum\limits_{k = 1}^K {{\alpha _k}({x_i},w)N({t_{ni}}|{\mu _k}({x_i},w),\sigma _k^2({x_i},w)} } \right\}} } \\
 - \ln \left\{ \begin{array}{l}
 \\
e({x_i})\textstyle \,\,\,\,\,\,\,\,\,\,\,\,\,\,\,\,\,\textrm{if}\,\,{x_i}\,\,\textrm{has}\,\,\textrm{any}\,\,\textrm{nucleus}\\
1 - e({x_i})\,\,\,\,\,\,\,\,\textrm{Otherwise}
\end{array} \right.
\end{array}
\end{equation}
where ${I}$ is the number of the training images, ${N_i}$ is the number of nucleus within each image and ${{t_{ni}}}$ is the coordinate of $n$th point within the image patch ${i}$. ${e_i}$ is a Bernoulli variable that specifies the probability of the patch containing any nucleus and the variance covering the dilation. 

\paragraph{\textbf{Pointset for each nuclei}:} 
During the training of the network, we use a dilated point set located within 6 pixels from the nucleus center. This augments the training data and increases the training efficiency. We sample 10 points from a Gaussian distribution with the mean on nucleus centroid.
\paragraph{\textbf{Network Architecture}:}
In this paper, a CNN model was selected because of its capability to deal directly with  raw images, without the need of preprocessing and an explicit features extraction process. The network is trained to capture the important aspects of the input data. By optimizing the dense representation of the input data in the feature maps, the performance of the fully connected part (MDN) is improved.

\begin{figure}[h] 
\centering
\includegraphics[width=1\textwidth]{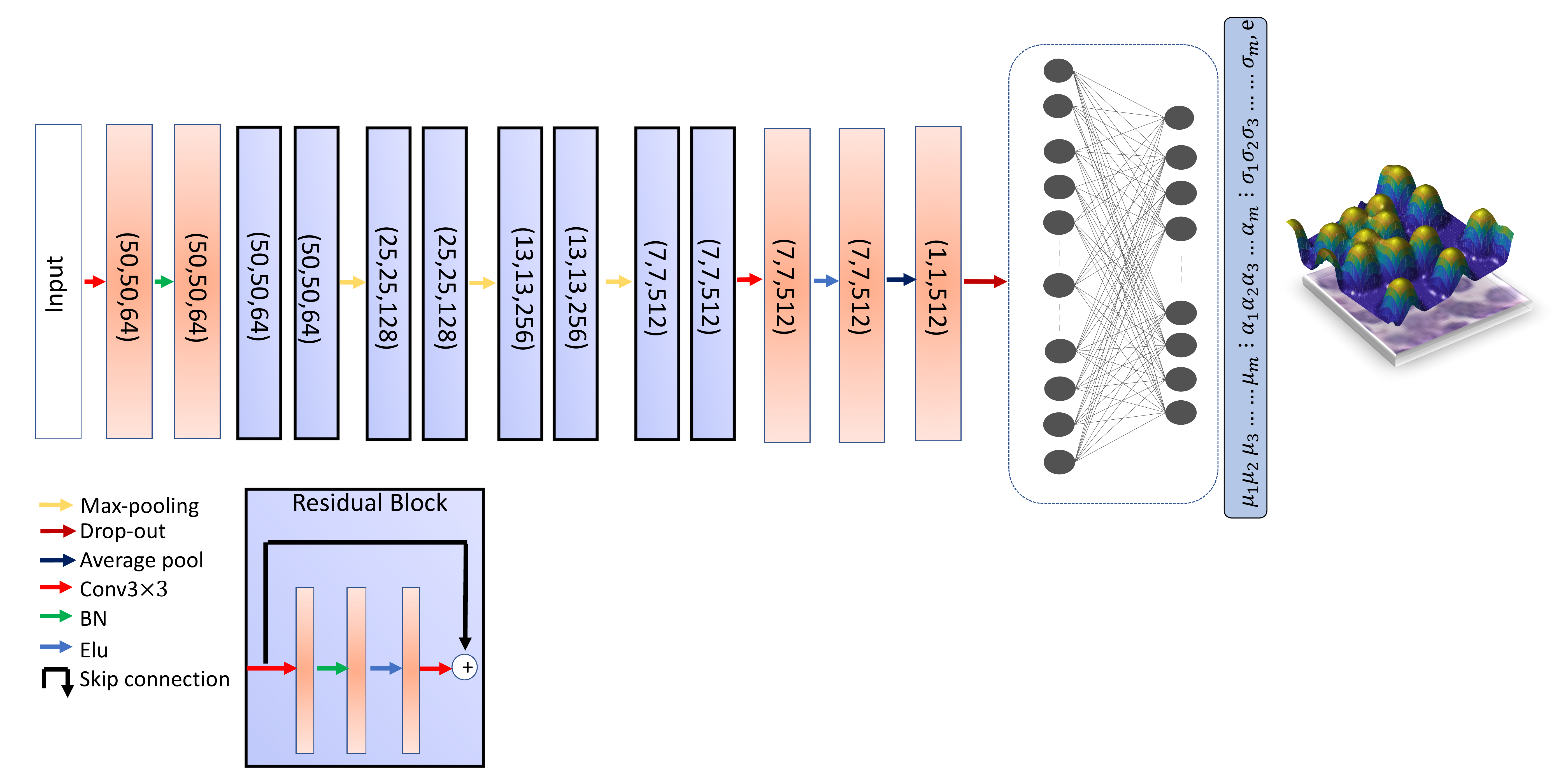}
\caption{The schematic architecture  of the proposed method.}
\label{schematic}
\end{figure}

For having a rich feature and better convergence, Resnet \cite{he2016deep} with 18 layers is utilized. In this architecture 'relu' activation function are replaced with 'elu'. We did not use very deep Resnet architecture as its training  requires huge amount of data. Two fully connected layers are added  after average-pooling to construct the whole architecture of MDN (See Fig \ref{schematic}).

To provide an appropriate input size to the network, the original images were cropped to patches of size 50$\times$50. The network architecture consists of 2 fully connected layers (256 and ${((c + 2) \times K)+1)}$, respectively). 
We set the number of mixtures to 100, therefore the MDN should predict 401 values (for each mixture 400 values and 1 value for the Bernoulli distribution). After acquiring network predictions, the patches with no nucleus having the low value of $e$ are ignored (threshold for $e$ is set to 0.5). We choose the most significant  Gaussians by  applying a threshold of 0.001 on the mixture coefficients (${\alpha _i}$). Afterward,  the probability maps are generated using ${\alpha _i}$,s, ${\sigma _i}$,s and ${\mu _i}$,s.  Finally to extract the centroids of the nuclei within the remaining patches, local maxima  are sought.

\section{Experimental Results} \label{Experimental}

\paragraph{\textbf{Dataset}:}
For our experiments, we use the  Colorectal cancer (CRC) dataset provided by \cite{sirinukunwattana2016locality}. It involves 100 H\&E images of colorectal adenocarcinomas of size 500$\times$500 which are cropped from CRC whole slide images.  The total number of 29756 nuclei were annotated for detection purpose. 
All the images are obtained at 20X magnification. This dataset is randomly divided
into two halves for training and testing. The cell detection on this
dataset is challenging due to touching cells, blurred (or weak) cell
boundaries and inhomogeneous background noise.

\begin{figure}[h]
	
	\centering
	\includegraphics[width=\textwidth]{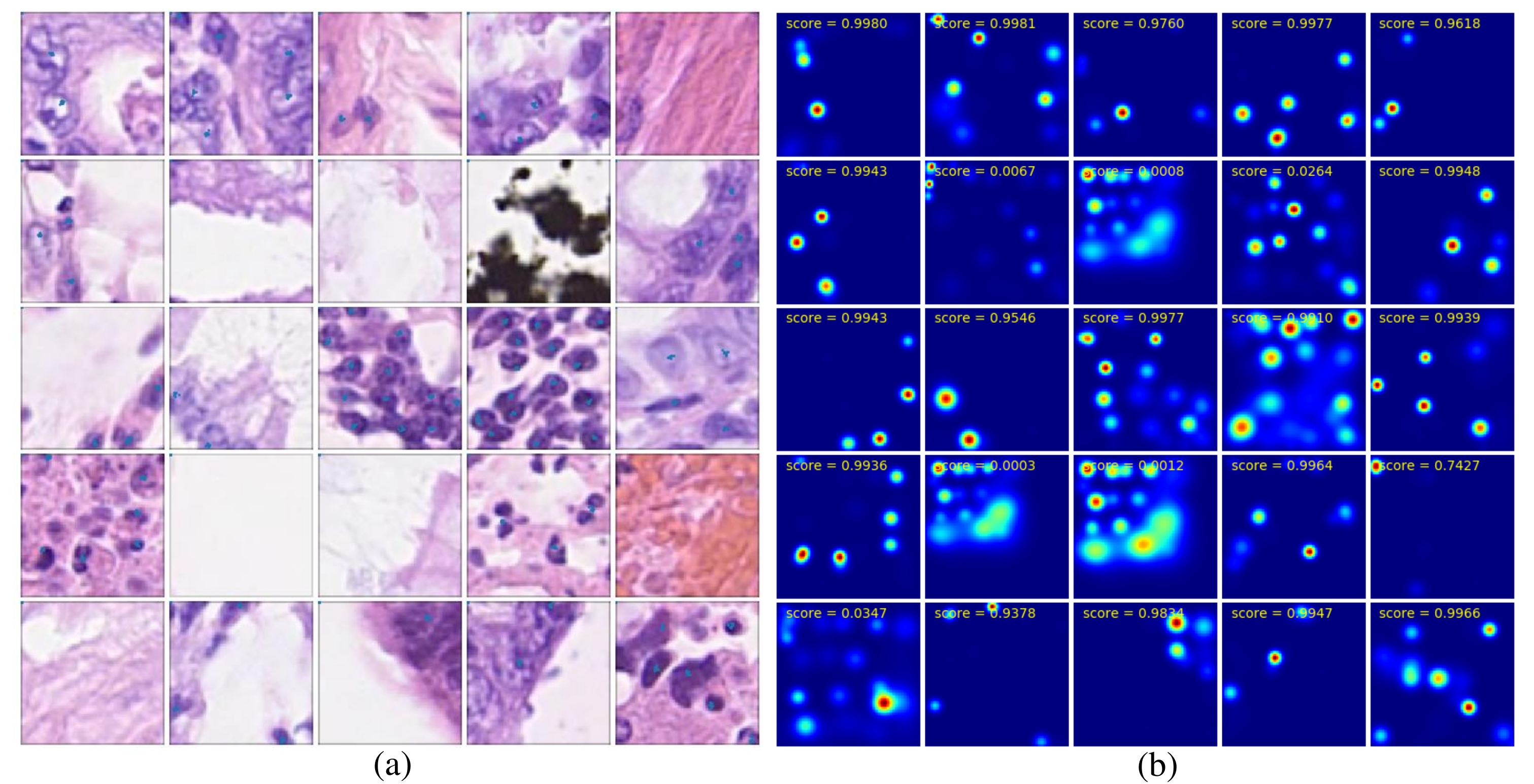}
	\caption{The image patches with their corresponding generated probability maps. a) The ground truth nuclei locations  is overlaid on the images. b)The corresponding probability map generated using our proposed MDN. The  score on top of each image is  showing the probability of that patch containing any nucleus. }
    \label{gg}
\end{figure}

\begin{figure}[h]
	\centering
	\includegraphics[width=\textwidth]{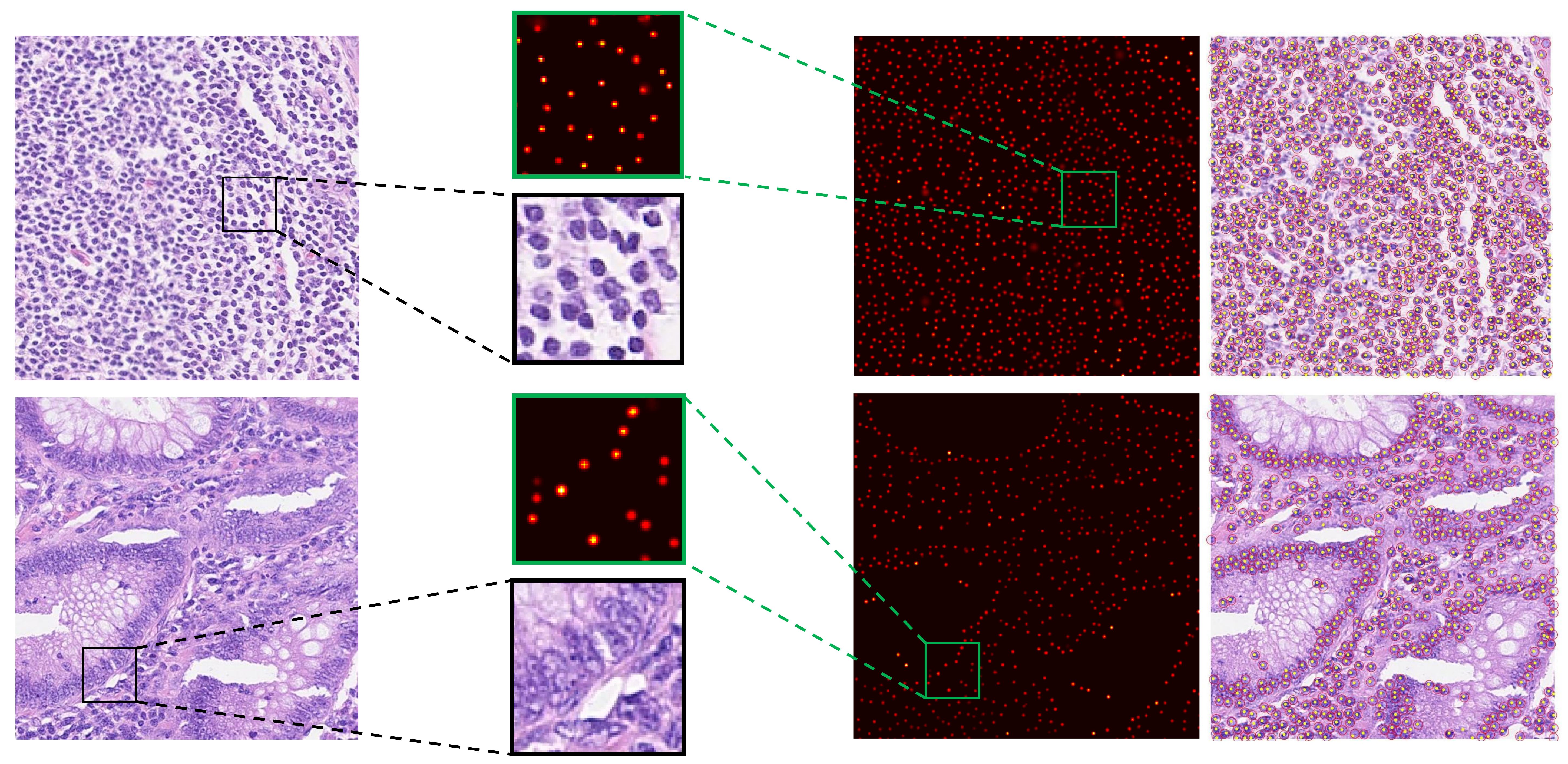}
	\caption{The original images on the left most column and their corresponding MDN outputs. For better visualization of congested lymphocyte nuclei (first row) and complex tumor epithelial,  regions of interest are enlarged in the green boxes. The right most column shows the ground truth specified  in yellow circles with detected nuclei as red dot. }
    \label{GT}
\end{figure}

\paragraph{\textbf{Results and Discussion}:} 
Fig. \ref{gg} shows the probability maps and the centroid locations along with the ground truth circles overlaid on the original images. As shown, the network could learn the locations of complex nuclei such as epithelial as well as congested area where lymphocyte nuclei lie. The broader view of the two challenging images and their corresponding probability maps are depicted in Figure \ref{GT}. For visual assessment, the annotated centroids (yellow circles) and predicted locations (red dots) are also shown in Fig \ref{GT}.  

\vspace{-6mm}

For the quantitative  evaluation we use the same two-fold cross validation explained in \cite{sirinukunwattana2016locality}.
Precision, Recall and F1 score are used for validating the detection performance. Each detected nucleus within the radius of 6 pixels from the annotated center is considered as true positive.  The final results are shown in table \ref{results}. The algorithm has low false negatives which leads to higher recall compared to other methods. In other word, high recall highlights its performance in detecting relatively more cells compared to its counterparts.   Overall the F1 score is high, which shows a good detection performance in the proposed MDN based framework.

Due to its probabilistic output, one advantage of the proposed method is its ability to handle images with weak and sparse annotations. We demonstrate this through the following procedure. Firstly we equally divide the dataset into training and validation sets and then  remove 30\% of the available annotations from the training set and compare the results with SR-CNN.  The quantitative results in Table \ref{results_sparse}, obtained using this sparsely annotated data, show that the proposed method can achieve a better performance. 
\begin{table}[h!]
	\caption{Comparison of precision, recall and F1 scores with other
		approaches.}
	\label{results}
	\centering
	\begin{tabular}{l|c|c|c}

		Method     & Precision     & Recall&F1 score \\\hline
		Proposed & 0.788& 0.882 & 0.832\\
		SC-CNN \cite{sirinukunwattana2016locality} & 0.781  & 0.823 & 0.802     \\
		SR-CNN \cite{xie2018efficient}     & 0.790      &0.834 &0.811  \\
		
		SSAE \cite{xu2016stacked} & 0.617 & 0.644 & 0.630 \\

	\end{tabular}
\end{table}
\begin{table}[h!]
	\caption{Comparison of precision, recall and F1 scores using weakly annotated data.}
	\label{results_sparse}
	\centering
	\begin{tabular}{l|c|c|c}
		Method     & Precision     & Recall&F1 score \\\hline
		Proposed & 0.67 & 0.75 & 0.71\\
		SR-CNN \cite{xie2018efficient}     & 0.59  & 0.63& 0.60  \\
	\end{tabular}
\end{table}

\section{Conclusion}\label{conclusion}
In this study, we used a probabilistic approach for detecting nucleus. MDN has been used in literature for one to many regression tasks. Here, we proposed a framework for employing MDN for nuclei detection. Firstly the features learned using a CNN taking images as input. Then, the MDN learns the distribution of nucleus within the image patch  using a mixture of Gaussian. Our method is capable of utilizing weak annotated data while preserving a good performance.  Finally, we showed that the proposed method can detect nucleus in colorectal histology images with a higher F1 score when compared to other approaches.

\bibliographystyle{splncs}
\bibliography{MyCollection}

\end{document}